\documentclass[10pt,a4paper]{article}
\newcommand{\BrandPrimaryHex}{2452D6}
\newcommand{\BrandTintHex}{F1F4F8}
\newcommand{\BrandTextHex}{000000}
\newcommand{\BrandMutedHex}{000000}
\newcommand{\BrandPanelHex}{2452D6}
\newcommand{\BrandName}{Symbiosis Robotics}
\newcommand{\BrandLogoPath}{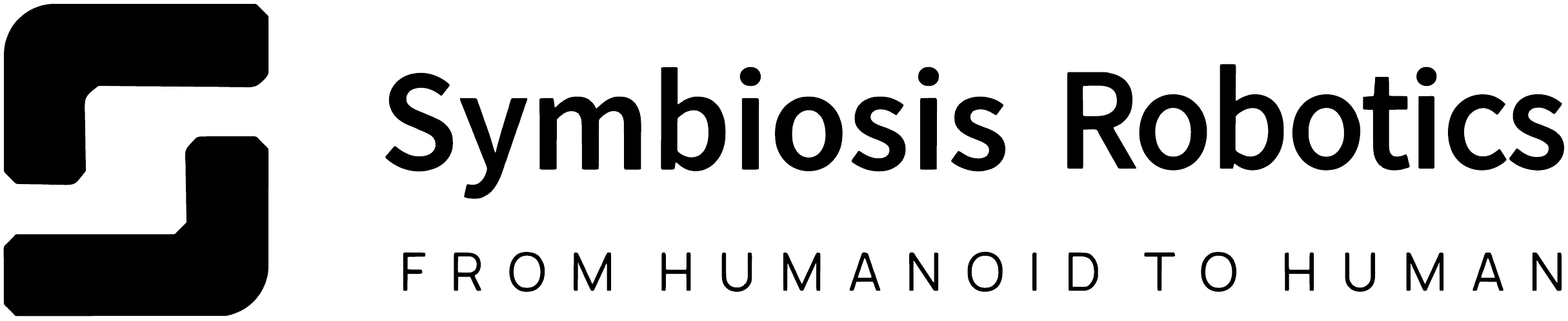}
\newcommand{\BrandLogoWidth}{42mm}
\newcommand{\BrandLogoHeight}{8.6mm}
\newcommand{\BrandMarkPath}{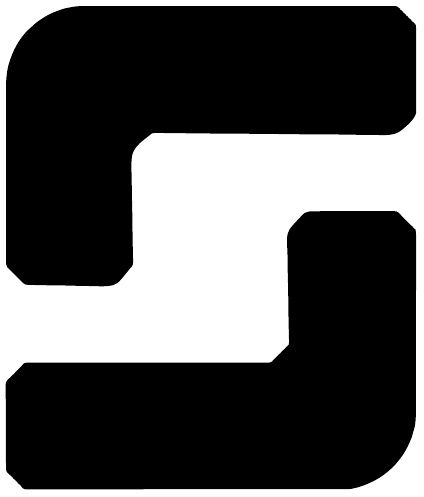}
\newcommand{\BrandMarkHeight}{4mm}

\usepackage{report-template}
\graphicspath{{figures/}}
\usepackage[section]{placeins}
\ifPDFTeX\pdfminorversion=7\fi

\newcommand{\ReportTitle}{ODEAction: From Predicting Chunks to Evolving Actions}
\newcommand{\ReportDisplayTitle}{ODEAction: From Predicting Chunks\\to Evolving Actions}
\newcommand{\ReportSubtitle}{}

\newcommand{\ReportAuthors}{%
  \mbox{Qishen Wang$^{1,2}$},
  \mbox{Pengxiang Ding$^{1,3,4\ddagger}$},
  \mbox{Shifeng Bao$^{5}$},
  \mbox{Mingyang Sun$^{1,3,4}$},
  \mbox{Hengtao Li$^{1}$},\\
  \mbox{Yi Gao$^{1}$},
  \mbox{Baohua Yin$^{1}$},
  \mbox{Yang Liu$^{4}$},
  \mbox{Jiajun Fu$^{1,6}$},
  \mbox{Jiacheng Liu$^{1,3,4}$},
  \mbox{Changjing He$^{1,6}$},\\
  \mbox{Peiyang Li$^{1,7}$},
  \mbox{Runze Suo$^{8}$},
  \mbox{Yan Wang$^{6}$},
  \mbox{Xiaobo Wang$^{9}$},
  \mbox{Yonghao Dang$^{10}$},
  \mbox{Cheng Chi$^{5*}$}}
\newcommand{\ReportAffiliations}{%
  \begingroup
  \fontfamily{SourceSerifPro-LF}\fontseries{l}\fontsize{9}{12}\selectfont\color{BrandText}
  \mbox{$^{1}$\,Symbiosis Robotics},
  \mbox{$^{2}$\,University of Science and Technology of China},
  \mbox{$^{3}$\,Zhejiang University},
  \mbox{$^{4}$\,Westlake University},
  \mbox{$^{5}$\,Renmin University of China},
  \mbox{$^{6}$\,Tsinghua University},
  \mbox{$^{7}$\,Beijing University of Posts and Telecommunications},
  \mbox{$^{8}$\,Fudan University},
  \mbox{$^{9}$\,Shenzhen University of Advanced Technology},
  \mbox{$^{10}$\,Central University of Finance and Economics}
  \par\vspace{3pt}%
  {\fontsize{8.5}{11}\selectfont$^{\ddagger}$\,Project Lead, $^{*}$\,Corresponding Author\par}%
  \endgroup}
\newcommand{\ReportPDFAuthors}{Qishen Wang, Pengxiang Ding, Shifeng Bao, Mingyang Sun, Hengtao Li, Yi Gao, Baohua Yin, Yang Liu, Jiajun Fu, Jiacheng Liu, Changjing He, Peiyang Li, Runze Suo, Yan Wang, Xiaobo Wang, Yonghao Dang, Cheng Chi}
\newcommand{\ReportType}{Technical Report}
\newcommand{\ReportVersion}{Working draft}
\newcommand{\ReportContact}{}

\begin{document}
\begin{reportfrontmatter}
\makereporttitle
\hypersetup{pdfsubject={\ReportType},pdfauthor={\ReportPDFAuthors}}
\begin{reportabstract}

Vision-language-action (VLA) models typically generate actions in chunks, dividing continuous robot execution into separately predicted action chunks, which can introduce discontinuities at chunk boundaries. We propose ODEAction, a chunk-free action generation paradigm that replaces action chunk prediction with Continuous Action Evolution. We use an ordinary differential equation (ODE) to describe the continuous evolution of actions, and define the vector field of this ODE as the Action Transition Field to characterize how actions evolve with respect to evolution time. During inference, the model takes the current action as the initial condition and numerically solves the ODE defined by the learned Action Transition Field, thereby recursively generating subsequent actions with explicit dependencies. ODEAction is implemented as a modular action head that can be integrated with different vision-language backbones. Experiments on LIBERO, EBench, RoboTwin, and real-robot tasks show that, compared with action-chunk-based baselines, ODEAction improves both task performance and trajectory quality, with particularly pronounced gains on fine manipulation tasks.

\end{reportabstract}

\end{reportfrontmatter}

\section{Introduction}

\label{sec:intro}

\begin{figure}[htbp]
  \centering
  \includegraphics[width=\linewidth]{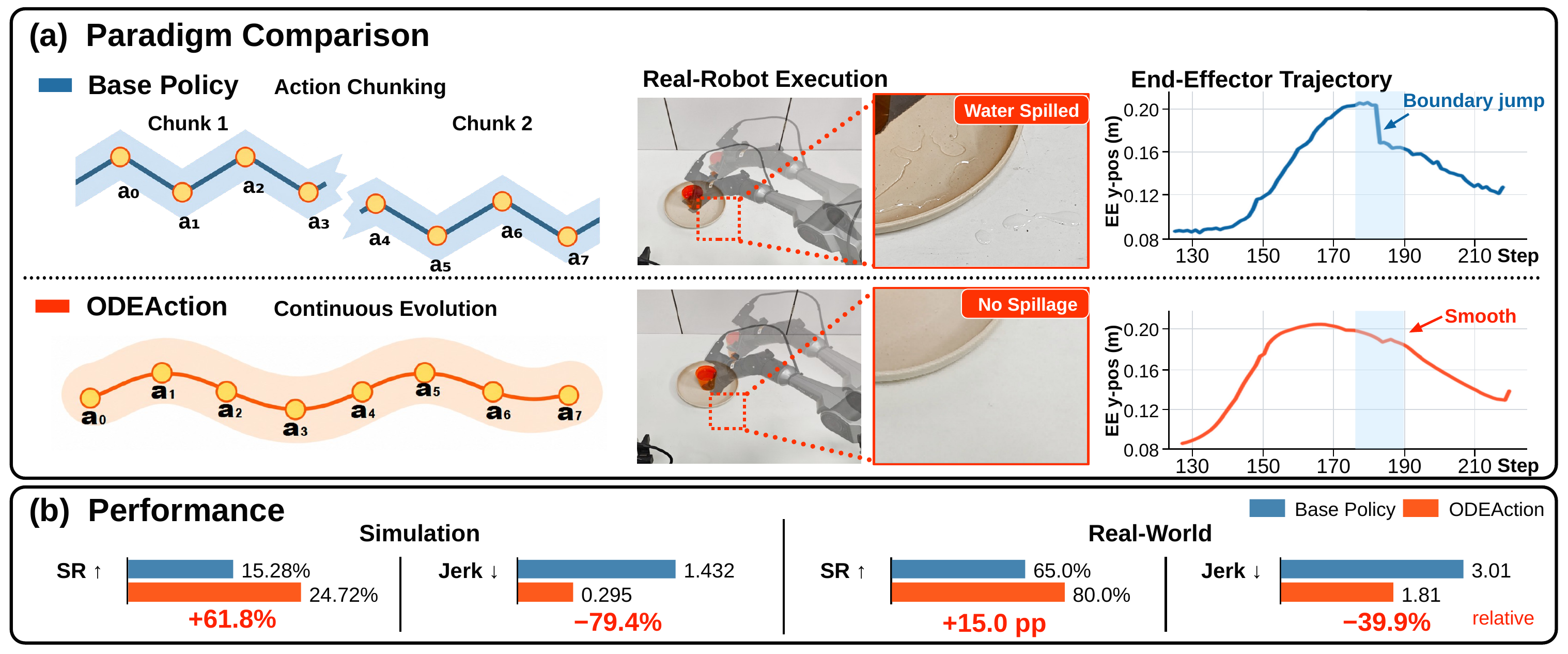}
  \caption{From action-chunk prediction to continuous action evolution. (a) Base Policy uses action-chunk prediction and is prone to abrupt action changes at chunk transitions, whereas ODEAction's continuous evolution ensures that action trajectories are generated continuously. Comparisons of real-robot task execution and the corresponding end-effector trajectories illustrate the differences in execution smoothness and trajectory continuity between the two methods. (b) Representative results on EBench simulation and real-robot tasks: compared with Base Policy, ODEAction increases the task success rate (SR) and reduces the end-effector mean absolute jerk (EE-MAJ).}
  \label{fig:overview}
\end{figure}

Vision-language-action (VLA) models~\cite{zitkovich2023rt,kim2024openvla,li2024roboflamingo,ReconVLA-2025,SpatialVLA-2025,RoboVLMs-2024,li2024cogact} widely adopt action-chunk generation (Action Chunking Generation): they jointly predict a sequence of future actions in a single inference pass, organizing continuous execution into successively generated action chunks~\cite{black2024pi_0,black2025pi_,bjorck2025gr00t,kim2025fine,shukor2025smolvla,PDVLA-2025,wen2025tinyvla}. However, adjacent chunks predicted independently typically lack explicit temporal consistency constraints across chunk boundaries, potentially causing abrupt action changes and trajectory discontinuities at transitions~\cite{black2025real,liu2024bidirectional} (\cref{fig:overview}(a)).

Existing methods alleviate this issue by imposing constraints on continuity across chunks: RTC~\cite{black2025real} constrains new action chunks during inference, whereas Training-time RTC~\cite{black2025training} and Legato~\cite{liu2026learning} introduce continuity constraints during training. Nevertheless, these methods still treat action chunks as the fundamental unit of generation, improving the connections between chunks rather than changing the action generation paradigm. This leads us to ask a further question: can we move beyond chunk-wise prediction and directly model the continuous evolution of actions over time?

Robot execution itself progresses incrementally over time rather than occurring in action chunks. We therefore characterize continuous action evolution using an ordinary differential equation (ODE): the current action provides the starting point, a learned action evolution rate determines subsequent changes, and integration progressively generates an action sequence without jointly predicting an entire future action chunk. Action generation thus shifts from predicting future action segments to learning the evolution of the current action.

From this perspective, we propose \textbf{ODEAction}, an action generation paradigm centered on \textbf{Continuous Action Evolution}. Given visual observations, language instructions, and proprioceptive information, ODEAction learns an Action Transition Field defined over action space. At each action and evolution time, the field outputs a transition vector representing the evolution rate of the current action with respect to evolution time. During training, we construct a target field around demonstration trajectories that combines progression along the demonstrations with deviation correction, and learn this field through direct supervision. During inference, starting from the current action, we integrate the learned action transition field using an ODE, so that each generated action becomes the starting point for subsequent evolution, instead of independently predicting and concatenating action chunks (\cref{fig:overview}(a)). ODEAction is implemented as an action head that can be integrated with different vision-language backbones.

We evaluate ODEAction on simulation benchmarks, including LIBERO~\cite{liu2023libero}, RoboTwin~\cite{chen2025robotwin}, and EBench~\cite{gao2026ebench}, as well as on real-robot tasks. Compared with matched action-chunking baselines, ODEAction improves both task performance and action trajectory quality, achieving a 61.8\% relative improvement in average success rate on EBench. Further analysis shows that evaluation units exhibiting larger improvements in boundary continuity also tend to achieve greater success-rate gains. In a representative execution of the real-world Place Cup task, ODEAction smoothly transports a water-filled cup without spilling, while producing a more continuous end-effector trajectory (\cref{fig:overview}(a)). Our contributions are threefold:

\begin{itemize}
  \item We propose ODEAction, a chunk-free generation paradigm based on continuous action evolution, transforming action generation from future action-chunk prediction into a learnable continuous evolution process.
  \item Extensive experiments on simulation benchmarks and real-robot tasks show that ODEAction improves both task performance and trajectory quality over matched action-chunking baselines, with particularly substantial benefits on fine manipulation tasks.
  \item We identify a systematic relationship between improvements in trajectory quality and gains in task performance: boundary discontinuity reduction is strongly positively correlated with success-rate gains, and evaluation units with greater improvements in trajectory continuity generally achieve larger task-performance gains.
\end{itemize}

\section{Method}

\label{sec:method}

\begin{figure}[htbp]
  \centering
  \includegraphics[width=\linewidth]{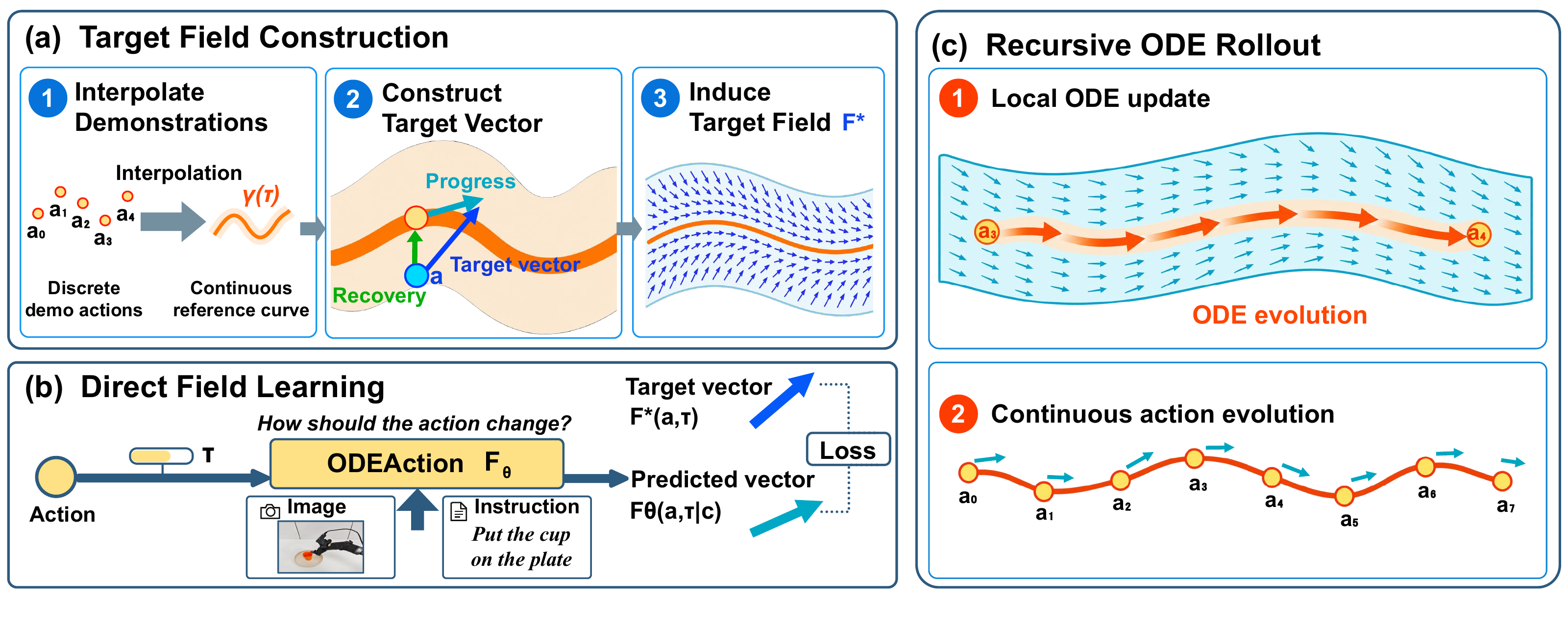}
  \caption{Learning the action transition field and recursively generating actions. (a) Discrete demonstrated actions are interpolated into a continuous trajectory. Target transition vectors are constructed by combining progression along the demonstration with deviation correction, inducing a target action transition field $F^*$ in the trajectory neighborhood. (b) During training, the action transition field ($F_\theta$) is learned through direct supervision. (c) During inference, the current action serves as the initial condition, and subsequent actions are recursively generated by numerically solving the ODE defined by the learned Action Transition Field.}
  \label{fig:method}
\end{figure}

ODEAction formulates action evolution as an ordinary differential equation (ODE) over the action space. As illustrated in \cref{fig:method}, we construct a demonstration-induced target field, learn the ODE's vector field through direct supervision, and numerically solve the learned ODE from the current action to recursively generate subsequent actions.

\subsection{Continuous Action Evolution}

\label{sec:continuous-evolution}

Unlike methods that jointly predict a future action chunk~\cite{zhao2023learning,chi2025diffusion,pertsch2025fast,black2024pi_0,kim2025fine}, ODEAction uses an ordinary differential equation to describe the continuous evolution of the current action over execution time. Let $c$ denote the conditioning information provided by visual observations, language instructions, and robot proprioceptive information, and let normalized evolution time $\tau\in[0,1]$ parameterize this process. Action evolution then satisfies

\begin{equation}
\frac{\mathrm d a(\tau)}{\mathrm d\tau}
=
F_\theta\bigl(a(\tau),\tau\mid c\bigr),
  \label{eq:continuous-evolution}
\end{equation}

where $a(\tau)$ denotes the action at evolution time $\tau$, and $F_\theta$ is a conditional vector field defined over action space, termed the \textbf{Action Transition Field}. Given an action and an evolution time, this field outputs a transition vector with the same dimensionality as the action, representing the action evolution rate of the current action with respect to evolution time.

\subsection{Demonstration-Induced Action Transition Field}

\label{sec:target-field}

Training demonstrations consist of discrete action sequences. We interpolate between adjacent demonstrated actions along evolution time to obtain a continuous demonstration trajectory $\gamma(\tau)$. Its derivative $\dot\gamma(\tau)=\mathrm d\gamma(\tau)/\mathrm d\tau$ gives the action evolution rate of the demonstration at evolution time $\tau$, providing supervision for the action transition field in \cref{eq:continuous-evolution} along the trajectory.

However, prediction errors or execution deviations during inference may cause the current action to drift away from the demonstration trajectory during recursive evolution. Supervising only the evolution rate on the trajectory cannot provide a corresponding evolution direction at off-trajectory positions. We therefore sample query actions $a=\gamma(\tau)+\sigma\epsilon$ around $\gamma(\tau)$, where $\epsilon\sim\mathcal N(0,I)$ and $\sigma$ controls the extent of neighborhood coverage. To enable off-trajectory actions both to progress along the demonstration and to return toward its path, we define the target action transition field as

\begin{equation}
F^{*}(a,\tau)
=
\dot{\gamma}(\tau)
+
k\bigl(\gamma(\tau)-a\bigr),
  \label{eq:target-field}
\end{equation}

Inspired by guiding-vector-field designs for robot path-following control~\cite{kapitanyuk2017guiding}, the first term specifies progression along the demonstration, while the second provides proportional correction toward the demonstration path for off-trajectory actions; $k$ controls the correction strength. When $a=\gamma(\tau)$, the correction term vanishes, and the target transition vector equals the demonstration's evolution rate. When the action deviates, the two terms jointly determine the direction of subsequent evolution. In this way, discrete demonstrations induce a target action transition field in the neighborhoods of their trajectories (\cref{fig:method}(a)). Supervision samples from different demonstrations, conditions, and evolution times are jointly used to learn the conditional action transition field $F_\theta$. ODEAction uses a fixed neighborhood scale $\sigma$ and correction gain $k$ throughout the entire evolution interval.

\subsection{Direct Supervision and ODE Rollout}

\label{sec:learning-rollout}

\textbf{Direct Supervision.} Given conditioning information $c$ and a demonstration trajectory $\gamma$ from the training data, we sample $\tau\sim\mathcal U[0,1]$, then sample a query action $a\sim\mathcal N(\gamma(\tau),\sigma^2I)$ in the neighborhood of the corresponding demonstrated action. We directly supervise $F_\theta$ using the target field $F^*$:

\begin{equation}
\mathcal L(\theta)
=
\mathbb E_{\substack{
(c,\gamma)\sim\mathcal D,\\
\tau\sim\mathcal U[0,1],\\
a\sim\mathcal N(\gamma(\tau),\sigma^2I)
}}
\left[
\left\|
F_\theta(a,\tau\mid c)
-
F^{*}(a,\tau)
\right\|_2^2
\right].
  \label{eq:training-objective}
\end{equation}

Each training sample provides a query action and its target transition vector, from which the model learns evolution rates under different conditions, action positions, and evolution times (\cref{fig:method}(b)). Training requires neither solving an ODE nor backpropagating through numerical integration.

\textbf{ODE Rollout.} During inference, we take the current action as the initial condition and numerically solve the ODE defined by the learned Action Transition Field, recursively generating subsequent actions through the resulting action evolution. We use the forward Euler method in all experiments. One integration step is given by

\begin{equation}
a_{n+1}
=
a_n
+
\Delta\tau\,
F_\theta(a_n,\tau_n\mid c).
  \label{eq:euler-step}
\end{equation}

\Cref{eq:euler-step} is the forward Euler discretization of \cref{eq:continuous-evolution}. Each generated action serves as the starting point for subsequent evolution, establishing explicit generative dependencies between adjacent actions (\cref{fig:method}(c)). When a new visual observation becomes available, the model updates the condition $c$ using the latest observation and continues integrating from the current action along the updated conditional action transition field. Each query to the action transition field outputs only one $d$-dimensional transition vector; subsequent actions are produced through recursive integration rather than through joint prediction of a fixed-length action chunk.

\section{Experiments}

\label{sec:experiments}

This section addresses the following questions:

\begin{itemize}
  \item Can ODEAction improve task performance and trajectory quality simultaneously? Are improvements in trajectory continuity associated with success-rate gains? (\cref{sec:chunking-comparison})
  \item Does ODEAction remain effective across different base policies and tasks? (\cref{sec:generality})
  \item Can ODEAction outperform methods designed to maintain continuity across action chunks? (\cref{sec:continuation-comparison})
  \item Which design choices are necessary for ODEAction's performance? (\cref{sec:ablation})
  \item Can ODEAction improve both task performance and trajectory quality on real-robot tasks? (\cref{sec:real-robot})
\end{itemize}

\subsection{Experimental Setup}

\label{sec:experimental-setup}

\textbf{Policies and Protocols.} We consider three base-policy configurations. With $\pi_{0.5}$~\cite{black2025pi_} as the base policy, we compare action-chunk generation and ODEAction on LIBERO and EBench, and additionally include RTC and Legato in the real-robot experiments. With StarVLA-OFT~\cite{community2026starvla,kim2025fine} as the base policy, we evaluate ODEAction on all 50 RoboTwin tasks. With StarVLA-$\pi$ as the base policy, we compare the original action-chunking method, RTC, Legato, and ODEAction on five representative RoboTwin tasks. LIBERO~\cite{liu2023libero}, RoboTwin~\cite{chen2025robotwin}, and EBench~\cite{gao2026ebench} cover tabletop manipulation, bimanual manipulation, and long-horizon fine manipulation, respectively. Real-robot experiments use AgileX Cobot Magic 1.0, and each benchmark follows its corresponding evaluation protocol. Within each comparison, we keep the base backbone, task-training data, control frequency, and visual-observation update times identical; details of evolution time and the execution mechanism are provided in Supplementary Materials~\ref{supp:section:a} and~\ref{supp:section:b}. Except for ablation experiments, ODEAction consistently uses $\sigma=0.1$ and $k=5$ across all benchmarks, tasks, and real-robot experiments, without task- or platform-specific adjustments; parameter selection is discussed in Supplementary Material~\ref{supp:section:a}.

\textbf{Metrics.} We measure task performance using the success rate (SR) across all evaluation episodes. For trajectory quality, we use four metrics, all of which are better when lower. NLDLJ, NSPARC~\cite{Balasubramanian2015OnTA,arachchige2025sail}, and EE-MAJ measure time-domain smoothness, frequency-domain smoothness, and end-effector mean absolute jerk, respectively, and are collectively referred to as Motion Smoothness metrics. Boundary Jump measures the magnitude of changes between adjacent control commands at chunk boundaries, characterizing Trajectory Continuity. For ODEAction, Boundary Jump is computed at the chunk-boundary times of the corresponding Base Policy. Trajectory metrics are calculated from model-output control commands in the successful episodes of each method; the separate evaluation protocol for offline recursive error is described in Supplementary Material~\ref{supp:section:c}.

\subsection{Comparison with Action Chunking}

\label{sec:chunking-comparison}

Using $\pi_{0.5}$ as the base policy, we compare task performance and trajectory quality between action-chunk generation and ODEAction on LIBERO and EBench, and analyze the relationship between improvements in boundary continuity and success-rate gains. To quantify improvements in boundary continuity relative to the action-chunking baseline, we define the Boundary Discontinuity Reduction (BDR) for evaluation unit $T$ as

\begin{equation}
\operatorname{BDR}(T)
=
\frac{
J^{\mathrm{bnd}}_{\mathrm{chunk}}(T)
-
J^{\mathrm{bnd}}_{\mathrm{ODEAction}}(T)
}{
J^{\mathrm{bnd}}_{\mathrm{chunk}}(T)
}.
  \label{eq:bdr}
\end{equation}

Here, $J^{\mathrm{bnd}}_{\mathrm{chunk}}$ and $J^{\mathrm{bnd}}_{\mathrm{ODEAction}}$ denote the Boundary Jump of Base Policy and ODEAction, respectively, both evaluated at the same Base Policy action-chunk boundary times. A larger BDR indicates a greater reduction in boundary discontinuity.

\textbf{Task Performance and Trajectory Quality.} As shown in \Cref{tab:main-results}, the average success rate on LIBERO increases from 96.85\% to 98.25\%, with improvements across all four suites. On EBench, it increases from 15.28\% to 24.72\%, representing a 61.8\% relative improvement. Across all evaluation units in both benchmarks, ODEAction simultaneously increases the success rate and improves all four trajectory metrics, with particularly pronounced performance gains on fine manipulation tasks.

\begin{table}[htbp]
  \caption{Comparison of task success rates and trajectory quality between ODEAction and Base Policy on LIBERO and EBench. SR gains are expressed in percentage points (pp), while improvements in NLDLJ, NSPARC, and EE-MAJ are relative reductions; EE-MAJ is reported in units of $10^{-3}$. The mean BDR is the arithmetic mean across evaluation units, and aggregate results are calculated from unrounded data.}
  \label{tab:main-results}
  \centering
  \setlength{\tabcolsep}{0pt}
  \resizebox{\textwidth}{!}{%
  \begin{tabular}{@{}l@{\hspace{10pt}}*{5}{c@{\hspace{4pt}}c@{\hspace{10pt}}}c@{}}
    \toprule
    & \multicolumn{2}{c@{\hspace{10pt}}}{SR (\%) $\uparrow$} & \multicolumn{2}{c@{\hspace{10pt}}}{NLDLJ $\downarrow$} & \multicolumn{2}{c@{\hspace{10pt}}}{NSPARC $\downarrow$} & \multicolumn{2}{c@{\hspace{10pt}}}{EE-MAJ $\downarrow$} & \multicolumn{2}{c@{\hspace{10pt}}}{Boundary Jump $\downarrow$} & BDR (\%) \\
    \cmidrule(l{-2pt}r{8pt}){2-3}\cmidrule(l{-2pt}r{8pt}){4-5}\cmidrule(l{-2pt}r{8pt}){6-7}\cmidrule(l{-2pt}r{8pt}){8-9}\cmidrule(l{-2pt}r{8pt}){10-11}
    Benchmark / Task & Base & Ours & Base & Ours & Base & Ours & Base & Ours & Base & Ours & \\
    \midrule
    \multicolumn{12}{l}{\textit{LIBERO}} \\
    Spatial & 98.8 & \textbf{99.3} & 26.35 & \textbf{25.32} & 2.121 & \textbf{1.983} & 1.19 & \textbf{0.48} & 0.04314 & \textbf{0.04283} & 0.72 \\
    Object & 98.2 & \textbf{99.8} & 25.87 & \textbf{24.46} & 2.104 & \textbf{2.010} & 1.14 & \textbf{0.39} & 0.04102 & \textbf{0.04040} & 1.51 \\
    Goal & 98.0 & \textbf{98.8} & 26.72 & \textbf{25.86} & 2.138 & \textbf{1.962} & 1.23 & \textbf{0.53} & 0.04577 & \textbf{0.04547} & 0.66 \\
    Long & 92.4 & \textbf{95.1} & 28.24 & \textbf{26.44} & 2.172 & \textbf{2.042} & 1.31 & \textbf{0.38} & 0.05213 & \textbf{0.05061} & 2.92 \\
    Average & 96.85 & \textbf{98.25} & 26.80 & \textbf{25.52} & 2.134 & \textbf{1.999} & 1.22 & \textbf{0.44} & 0.04552 & \textbf{0.04483} & 1.45 \\
    \textit{Relative improvement} & \multicolumn{2}{c@{\hspace{10pt}}}{\textit{+1.40 pp}} & \multicolumn{2}{c@{\hspace{10pt}}}{\textit{4.76\%}} & \multicolumn{2}{c@{\hspace{10pt}}}{\textit{6.31\%}} & \multicolumn{2}{c@{\hspace{10pt}}}{\textit{63.70\%}} & \multicolumn{2}{c@{\hspace{10pt}}}{---} & --- \\
    \midrule
    \multicolumn{12}{l}{\textit{EBench}} \\
    Coffee & 5.0 & \textbf{13.3} & 27.88 & \textbf{26.24} & 2.138 & \textbf{2.025} & 1.38 & \textbf{0.29} & 0.01680 & \textbf{0.01390} & 17.26 \\
    Flip Cup & 5.0 & \textbf{10.0} & 29.15 & \textbf{26.73} & 2.118 & \textbf{2.034} & 1.62 & \textbf{0.34} & 0.02000 & \textbf{0.01810} & 9.50 \\
    Frame & 30.0 & \textbf{48.3} & 26.70 & \textbf{25.65} & 2.122 & \textbf{2.022} & 1.53 & \textbf{0.31} & 0.01520 & \textbf{0.01140} & 25.00 \\
    Gear & 15.0 & \textbf{23.3} & 24.36 & \textbf{22.91} & 2.125 & \textbf{2.014} & 1.50 & \textbf{0.29} & 0.01400 & \textbf{0.01300} & 7.14 \\
    Glass & 25.0 & \textbf{30.0} & 28.55 & \textbf{26.07} & 2.121 & \textbf{2.027} & 1.11 & \textbf{0.26} & 0.02450 & \textbf{0.02290} & 6.53 \\
    Nut & 11.7 & \textbf{23.3} & 26.29 & \textbf{25.47} & 2.145 & \textbf{2.016} & 1.46 & \textbf{0.28} & 0.01180 & \textbf{0.00990} & 16.10 \\
    Average & 15.28 & \textbf{24.72} & 27.16 & \textbf{25.51} & 2.128 & \textbf{2.023} & 1.43 & \textbf{0.30} & 0.01705 & \textbf{0.01487} & 13.59 \\
    \textit{Relative improvement} & \multicolumn{2}{c@{\hspace{10pt}}}{\textit{+9.44 pp}} & \multicolumn{2}{c@{\hspace{10pt}}}{\textit{6.08\%}} & \multicolumn{2}{c@{\hspace{10pt}}}{\textit{4.93\%}} & \multicolumn{2}{c@{\hspace{10pt}}}{\textit{79.40\%}} & \multicolumn{2}{c@{\hspace{10pt}}}{---} & --- \\
    \bottomrule
  \end{tabular}
  }
\end{table}

\begin{figure}[htbp]
  \centering
  \includegraphics[width=0.60\linewidth]{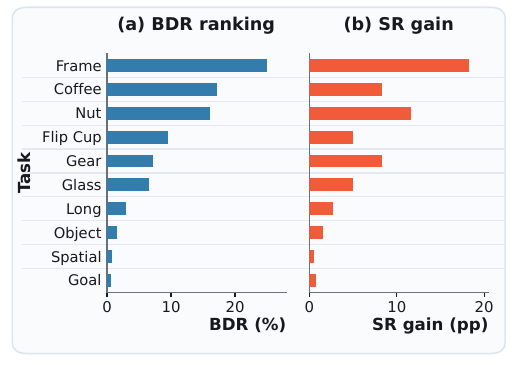}
  \caption{Relationship between improvements in boundary continuity and gains in task performance. (a) Four LIBERO suites and six EBench tasks arranged in descending order of Boundary Discontinuity Reduction (BDR). (b) Success-rate gains (SR gain) of ODEAction relative to Base Policy, shown in the same evaluation-unit order. BDR and success-rate gains exhibit a strong positive correlation across the ten evaluation units.}
  \label{fig:bdr}
\end{figure}

\textbf{Association Between BDR and Task-Performance Gains.} The average BDR values on LIBERO and EBench are 1.45\% and 13.59\%, respectively, corresponding to average success-rate gains of +1.40 and +9.44 percentage points. As shown in \cref{fig:bdr}(a), when ranked in descending order of BDR, all six EBench tasks precede the four LIBERO suites; \cref{fig:bdr}(b) shows the corresponding success-rate gains for each evaluation unit in the same order. Across all ten evaluation units, BDR is strongly positively correlated with success-rate gains (Spearman $\rho_s=0.939$; detailed statistics are provided in Supplementary Material~\ref{supp:section:d}). This indicates that evaluation units with greater improvements in boundary continuity also tend to achieve larger success-rate gains.

\subsection{Generality of the Generation Paradigm}

\label{sec:generality}

To examine generality across policies and tasks, we use StarVLA-OFT as the Base Policy and evaluate ODEAction on all 50 RoboTwin tasks. As shown in \Cref{tab:generality}, ODEAction increases the average success rate across the 50 RoboTwin tasks from 85.7\%/87.3\% to 89.4\%/89.8\% under the Clean/Random settings, respectively (per-task results are provided in Supplementary Material~\ref{supp:section:e}). All four control-command trajectory metrics also improve: NLDLJ decreases from 25.96 to 23.89, NSPARC from 2.298 to 2.159, EE-MAJ from 1.145 to 0.635, and Boundary Jump from 0.11200 to 0.08710. Together with the findings in \cref{sec:chunking-comparison}, these results show that ODEAction's benefits for task performance and trajectory quality extend across base policies and benchmarks.

\begin{table}[htbp]
  \caption{Success rates (\%) on all 50 RoboTwin tasks. Base Policy denotes StarVLA-OFT~\cite{community2026starvla}.}
  \label{tab:generality}
  \centering
  \setlength{\tabcolsep}{5pt}
  \begin{tabular}{lcc}
    \toprule
    Method & Clean $\uparrow$ & Random $\uparrow$ \\
    \midrule
    $\pi_0$~\cite{black2024pi_0} & 65.9 & 58.4 \\
    $\pi_{0.5}$~\cite{black2025pi_} & 82.7 & 76.8 \\
    ABot-M0~\cite{yang2026abot} & 86.1 & 85.1 \\
    X-VLA~\cite{zheng2026x} & 72.9 & 72.8 \\
    LingBot-VLA~\cite{wu2026pragmatic} & 88.6 & 86.7 \\
    \rowcolor{black!6} Base Policy~\cite{yuan2026qwen} & 85.7 & 87.3 \\
    \rowcolor{black!6} ODEAction (Ours) & \textbf{89.4} & \textbf{89.8} \\
    \bottomrule
  \end{tabular}
\end{table}

\subsection{Comparison with Action-Chunk Continuation Methods}

\label{sec:continuation-comparison}

\textbf{Comparison Setup.} To compare action-chunk continuation with continuous action evolution, we use StarVLA-$\pi$ as the Base Policy and establish four comparison groups on five RoboTwin tasks featuring different motion constraints and bimanual coordination patterns: Base Policy uses the original action-chunking method; Base Policy + RTC~\cite{black2025real} constrains new action chunks using already determined actions during inference; Base Policy + Legato~\cite{liu2026learning} incorporates known-action conditions during training to learn cross-chunk continuation; and Base Policy + ODEAction replaces the action head with ODEAction. All four groups share the same vision-language backbone, training data, control frequency, and observation-update times. Task and implementation details are provided in Supplementary Material~\ref{supp:section:b}.

\begin{table}[htbp]
  \caption{Comparison of four methods on five RoboTwin tasks, using StarVLA-$\pi$ as the Base Policy. EE-MAJ is reported in units of $10^{-3}$.}
  \label{tab:continuation}
  \centering
  \setlength{\tabcolsep}{0pt}
  \resizebox{\textwidth}{!}{%
  \begin{tabular}{@{}l@{\hspace{8pt}}*{4}{*{3}{c@{\hspace{3pt}}}c@{\hspace{8pt}}}*{3}{c@{\hspace{3pt}}}c@{}}
    \toprule
    & \multicolumn{4}{c@{\hspace{8pt}}}{SR (\%) $\uparrow$} & \multicolumn{4}{c@{\hspace{8pt}}}{NLDLJ $\downarrow$} & \multicolumn{4}{c@{\hspace{8pt}}}{NSPARC $\downarrow$} & \multicolumn{4}{c@{\hspace{8pt}}}{EE-MAJ $\downarrow$} & \multicolumn{4}{c}{Boundary Jump $\downarrow$} \\
    \cmidrule(l{-1.5pt}r{6.5pt}){2-5}\cmidrule(l{-1.5pt}r{6.5pt}){6-9}\cmidrule(l{-1.5pt}r{6.5pt}){10-13}\cmidrule(l{-1.5pt}r{6.5pt}){14-17}\cmidrule(l{-1.5pt}r{0pt}){18-21}
    Task & \small Base & \small +RTC & \small +Legato & \small Ours & \small Base & \small +RTC & \small +Legato & \small Ours & \small Base & \small +RTC & \small +Legato & \small Ours & \small Base & \small +RTC & \small +Legato & \small Ours & \small Base & \small +RTC & \small +Legato & \small Ours \\
    \midrule
    Adjust Bottle & 92 & 91 & 93 & \textbf{94} & 23.68 & 23.66 & 23.71 & \textbf{21.54} & 1.998 & 2.030 & 2.051 & \textbf{1.991} & 1.095 & 1.068 & 1.020 & \textbf{0.625} & 0.1720 & 0.1562 & 0.1616 & \textbf{0.1507} \\
    Handover Block & 79 & 80 & 75 & \textbf{82} & 29.10 & 28.17 & 27.59 & \textbf{26.67} & 2.850 & 2.954 & 2.868 & \textbf{2.735} & 1.248 & 1.050 & 0.833 & \textbf{0.581} & 0.0677 & 0.0630 & 0.0612 & \textbf{0.0391} \\
    Open Laptop & 85 & 82 & 78 & \textbf{87} & 25.53 & 25.15 & 25.30 & \textbf{22.73} & 2.023 & 2.009 & 1.996 & \textbf{1.848} & 1.037 & 0.987 & 1.046 & \textbf{0.449} & 0.0781 & 0.0681 & 0.0660 & \textbf{0.0264} \\
    Lift Pot & 88 & 90 & 86 & \textbf{91} & 23.34 & 22.64 & 22.59 & \textbf{20.69} & 1.723 & 1.681 & 1.698 & \textbf{1.566} & 1.170 & 0.817 & 0.724 & \textbf{0.511} & 0.1904 & 0.1712 & 0.1725 & \textbf{0.1613} \\
    Place Can Basket & \textbf{68} & 63 & 58 & \textbf{68} & 25.19 & 24.63 & 24.57 & \textbf{23.18} & 2.554 & 2.502 & 2.550 & \textbf{2.338} & 0.848 & 0.709 & 0.662 & \textbf{0.500} & 0.0742 & 0.0540 & 0.0497 & \textbf{0.0462} \\
    \midrule
    Average & 82.4 & 81.2 & 78.0 & \textbf{84.4} & 25.37 & 24.85 & 24.75 & \textbf{22.96} & 2.230 & 2.235 & 2.233 & \textbf{2.096} & 1.080 & 0.926 & 0.857 & \textbf{0.533} & 0.1165 & 0.1025 & 0.1022 & \textbf{0.0847} \\
    \bottomrule
  \end{tabular}
  }
\end{table}

\textbf{Results.} As shown in \Cref{tab:continuation}, the average success rates of Base Policy, +RTC, +Legato, and +ODEAction are 82.4\%, 81.2\%, 78.0\%, and 84.4\%, respectively. Although RTC and Legato reduce Boundary Jump from 0.1165 to 0.1025 and 0.1022, their success rates decrease by 1.20 and 4.40 percentage points, respectively. In contrast, ODEAction increases the success rate to 84.4\% while reducing Boundary Jump to 0.0847 (a 27.3\% reduction). Across the five tasks, ODEAction achieves the sole highest success rate on four tasks and ties for the highest on one; it also achieves the best values for all four trajectory-quality metrics on every task. These results demonstrate simultaneous improvements in task performance, motion smoothness, and trajectory continuity through continuous action evolution.

\subsection{Ablation Studies}

\label{sec:ablation}

\textbf{Ablation Setup.} Two key design choices jointly shape ODEAction's target action transition field: the action neighborhood scale $\sigma$ determines the extent of supervision coverage around demonstration trajectories, while the correction gain $k$ determines how strongly deviating actions are driven back toward the demonstration evolution path. To validate the roles of these two choices, we conduct ablations using the base policy and five tasks from \cref{sec:continuation-comparison}. We compare the full ODEAction with two variants: w/o neighborhood sets $\sigma=0$, providing supervision only on the demonstration trajectory; w/o correction sets $k=0$, retaining neighborhood sampling but removing the correction term. Because the correction term naturally vanishes when $\sigma=0$, these three groups cover all effective ablation combinations of the two designs. We additionally include Direct Next-Action, which likewise removes action chunks, to examine whether ODEAction's performance advantage stems solely from action-by-action recursion.

\begin{table}[htbp]
  \caption{Ablation of key ODEAction design choices. The w/o neighborhood variant sets $\sigma=0$, and the w/o correction variant sets $k=0$. $E$ denotes the full-trajectory offline rollout error, while $E_1$ and $E_2$ denote errors over the first and second halves, respectively; all are reported in units of $10^{-4}$. $E_2/E_1$ is the error accumulation ratio.}
  \label{tab:ablation}
  \centering
  \setlength{\tabcolsep}{3.5pt}
  \begin{tabular}{lccccc}
    \toprule
    Method & SR (\%) $\uparrow$ & $E$ $\downarrow$ & $E_1$ $\downarrow$ & $E_2$ $\downarrow$ & $E_2/E_1$ $\downarrow$ \\
    \midrule
    w/o neighborhood & 47.4 & 8.84 & 5.78 & 11.90 & 2.06 \\
    w/o correction & 56.2 & 13.17 & 8.26 & 18.08 & 2.19 \\
    ODEAction & \textbf{84.4} & \textbf{4.25} & \textbf{4.42} & \textbf{4.07} & \textbf{0.92} \\
    \bottomrule
  \end{tabular}
\end{table}

\textbf{Metrics.} In addition to success rate, we perform an offline rollout starting from the beginning of a demonstration trajectory under fixed conditions and compare the recursively generated actions with the demonstrated actions at corresponding evolution times, obtaining error $E$. Denoting the errors over the first and second halves of the rollout by $E_1$ and $E_2$, we define the Error Accumulation Ratio as $E_2/E_1$. A smaller ratio indicates weaker relative error growth in the second half, while a ratio greater than 1 indicates a larger error in the second half. The detailed calculation is provided in Supplementary Material~\ref{supp:section:c}.

\textbf{Both the Action Neighborhood and Correction Term Are Necessary.} As shown in \Cref{tab:ablation}, removing the neighborhood or correction term reduces the success rate from 84.4\% to 47.4\% or 56.2\%, respectively, while increasing offline error from 4.25 to 8.84 or 13.17. The error accumulation ratios of the two ablated variants are 2.06 and 2.19, compared with only 0.92 for the full ODEAction. Neighborhood supervision covers off-trajectory actions, while the correction term provides a direction for returning to the demonstration path. Together, they suppress the sustained growth of recursive errors.

\textbf{Distinction from Ordinary Action-by-Action Recursion.} Direct Next-Action recursively generates actions by directly predicting the next action. Under the same five-task setting, it achieves an average success rate of 78.4\%, below ODEAction's 84.4\%, and performs worse on all four trajectory-quality metrics (Supplementary Material~\ref{supp:section:f}). This result indicates that ODEAction's benefits do not arise solely from removing action chunks, but from the combined effects of modeling continuous action evolution rates and supervising the demonstration-neighborhood target field.

\subsection{Real-Robot Experiments}

\label{sec:real-robot}

\begin{figure}[htbp]
  \centering
  \includegraphics[width=\linewidth]{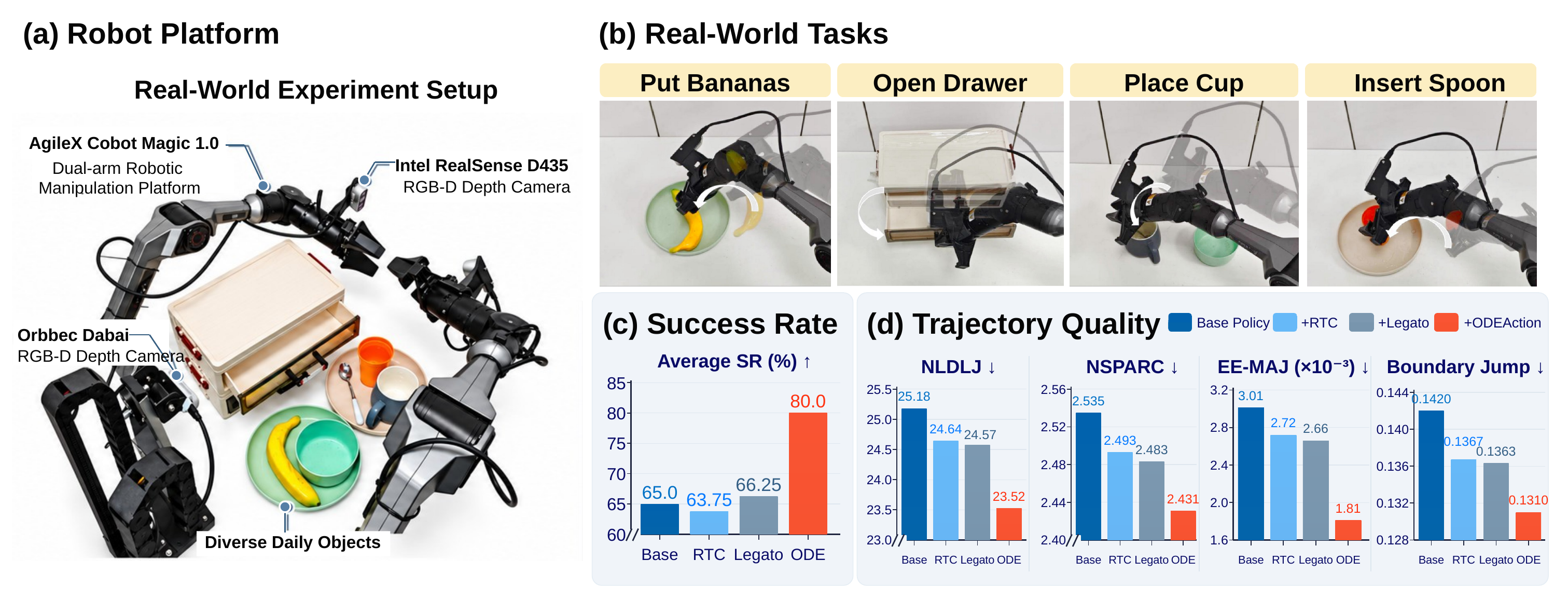}
  \caption{Real-robot experimental platform, tasks, and quantitative evaluation. (a) AgileX Cobot Magic 1.0 dual-arm robotic platform and visual sensor configuration. (b) Real-robot tasks. (c) Comparison of average success rates (SR) across the four tasks. (d) Comparison of average trajectory quality across the four tasks.}
  \label{fig:real-tasks}
\end{figure}

\textbf{Setup.} On the AgileX Cobot Magic 1.0 platform (\cref{fig:real-tasks}(a)), we use $\pi_{0.5}$ as the Base Policy and compare the original action-chunking method, RTC, Legato, and ODEAction. The four real-robot tasks (\cref{fig:real-tasks}(b)) are Put Bananas, Open Drawer, Place Cup, and Insert Spoon, covering pick-and-place, articulated-object manipulation, placement of a water-filled cup, and precise insertion. Place Cup requires that no water be spilled throughout execution. We collect 100 demonstration trajectories for each task and test each method 20 times per task, calculating success rates from the number of successful trials.

\textbf{Results.} As shown in \cref{fig:real-tasks}(c), Base Policy, +RTC, +Legato, and +ODEAction achieve average success rates of 65.00\%, 63.75\%, 66.25\%, and 80.00\%, respectively. ODEAction improves upon Base Policy by 15.00 percentage points and also outperforms RTC and Legato. In terms of trajectory quality, ODEAction outperforms the other methods on all four control-command trajectory metrics in each of the four tasks (complete per-task results are provided in Supplementary Table~\ref{supp:tab:s2}). The averaged results are shown in \cref{fig:real-tasks}(d): EE-MAJ decreases from 3.01 to 1.81, and Boundary Jump decreases from 0.14200 to 0.13100. Gains are particularly pronounced in fine manipulation: the success rate on Place Cup rises from 70.0\% to 90.0\% (see \cref{fig:overview}(a) for execution and trajectory comparisons), and that on Insert Spoon rises from 25.0\% to 55.0\%.

\section{Related Work}
\label{sec:related-work}

\subsection{Action Chunking and Cross-Chunk Continuation}
\label{sec:related-vla}
\label{sec:related-continuation}

Action chunking~\cite{zhao2023learning,chi2025diffusion} has become a common approach to action generation in large robot foundation models~\cite{team2024octo,liu2024rdt,team2025gemini,cheang2025gr,Zhao2025CoTVLAVC,Wang2025VQVLAIV,qu2025eo,kim2025fine,black2024pi_0,li2024cogact}, with models such as GR00T N1~\cite{bjorck2025gr00t} and $\pi_{0.5}$~\cite{black2025pi_} jointly generating a sequence of future actions in a single pass. However, independently predicting adjacent action chunks can lead to action discontinuities at chunk transitions. To improve continuity across chunks, RTC~\cite{black2025real} constrains new chunk generation using the overlapping portion of the previous action chunk during inference, while other methods enhance cross-chunk consistency through candidate selection~\cite{liu2024bidirectional} or generation guidance~\cite{arachchige2025sail,malhotra2025self}. Training-time RTC~\cite{black2025training} and Legato~\cite{liu2026learning} incorporate known actions as conditioning information during training, enabling policies to learn chunk continuation; related studies also combine masked action learning with prefix-preserved sampling to improve action consistency during asynchronous execution~\cite{wang2026real}. These methods retain action chunks as the unit of generation, whereas ODEAction recursively generates subsequent actions through continuous action evolution.

\subsection{Dynamical Systems and Continuous Action Generation}
\label{sec:related-paradigm}

Continuous dynamical systems have a long history in modeling robot motion. Prior work uses movement primitives and stable motion fields to reproduce demonstrated behaviors~\cite{ijspeert2013dynamical,khansari2011learning}. In robot path-following control, guiding vector fields combine tangential guidance with correction toward a reference path~\cite{kapitanyuk2017guiding}, while related robot-motion research develops stable dynamical systems for demonstrated motions~\cite{li2025movement} and progression--attraction flows for policy imitation~\cite{li2026geometry}. This perspective has also been incorporated into visuomotor policies: NDP~\cite{bahl2020neural} and related approaches embed movement primitives into neural policies~\cite{bahl2021hierarchical}, while other methods explore streaming action generation through continuous dynamics in action space~\cite{jiang2026guided,long2026constraining}. Under vision-language conditioning, studies have further explored the continuous evolution of proprioceptive representations~\cite{qi2026continuous} and the integration of vision-language reasoning with movement primitive generation~\cite{dai2026language,anarossi2025keymps}. SFP~\cite{jiang2025streaming} applies a progression-and-correction structure analogous to the guiding vector fields used in robot path-following control~\cite{kapitanyuk2017guiding} to streaming visuomotor imitation, formulating action generation as flow matching and incrementally generating actions by transporting an initial action distribution along a demonstration-conditioned flow. ODE Policy~\cite{jia2026ode} likewise models action generation through continuous action dynamics, but computes its training loss on trajectories generated by numerically solving the ODE, whereas ODEAction directly supervises the action transition field without performing ODE integration during training. ODEAction enables a policy to learn how the action itself evolves rather than predict future action chunks, modeling action evolution as an ordinary differential equation (ODE) in action space and recursively generating subsequent actions by numerically solving the ODE.

\section{Conclusion}

\label{sec:conclusion}

We propose ODEAction, a chunk-free action generation paradigm that replaces future action-chunk prediction with Continuous Action Evolution. Conditioned on visual, language, and proprioceptive information, ODEAction learns the vector field of an action-space ODE and recursively generates subsequent actions by numerically solving the ODE from the current action. Experiments on LIBERO, EBench, RoboTwin, and real robots show that ODEAction improves both task performance and trajectory quality over action-chunking baselines, with particularly pronounced benefits on fine manipulation tasks. We further find a strong positive correlation between boundary discontinuity reduction and success-rate gains: evaluation units with larger improvements in trajectory continuity also tend to achieve greater success-rate gains. Overall, ODEAction offers a new modeling paradigm for action generation in VLAs and robot control.

% Shared bibliography for the main text and appendix.
\bibliographystyle{plainnat}
\bibliography{references}
% Supplementary material supplied as English Markdown; see source/ for provenance.
\clearpage
\appendix
\setcounter{equation}{0}
\setcounter{figure}{0}
\setcounter{table}{0}
\renewcommand{\theequation}{S\arabic{equation}}
\renewcommand{\thefigure}{S\arabic{figure}}
\renewcommand{\thetable}{S\arabic{table}}
% Distinct hyperlink destinations from main-text equations, figures and tables.
\renewcommand{\theHequation}{supp.\arabic{equation}}
\renewcommand{\theHfigure}{supp.\arabic{figure}}
\renewcommand{\theHtable}{supp.\arabic{table}}
\section*{Supplementary Material}
\section{Additional Implementation Details of ODEAction}

\label{supp:section:a}

This section supplements the implementation details in Sec.~\ref{sec:method} of the main paper, describing the action head's inputs and outputs, the construction of demonstration supervision, the discretization of evolution time, integration and execution, and global parameters. ODEAction's continuous evolution, target field, and loss are defined in Eqs.~(\ref{eq:continuous-evolution})--(\ref{eq:training-objective}) of the main paper, respectively, and the Euler recursion is given in Eq.~(\ref{eq:euler-step}) of the main paper.

\textbf{Action Head Inputs and Outputs.} Given visual observations $o$, a language instruction $l$, and robot proprioceptive information $s$, the vision-language backbone produces a conditioning representation $c=\phi(o,l,s)$. The ODEAction action head takes the current action $a$, evolution time $\tau$, and condition $c$ as inputs and outputs a transition vector $F_\theta(a,\tau\mid c)\in\mathbb R^d$ with the same dimensionality as the action. This vector represents the rate of change of the action with respect to evolution time, rather than the next action itself or a fixed-length action chunk. When used with the base policies in Sec.~\ref{sec:experiments} of the main paper, the action representation, visual conditioning, and low-level control interface follow the configuration of the corresponding matched comparison group.

\textbf{Evolution Time, Interpolation, and Neighborhood Supervision.} Using control steps as discrete time units, let $t_o$ denote the initial control step aligned with the current conditioning representation $c$. An action evolution window spans $T_w$ control steps, giving

\begin{equation}
\tau=\frac{t-t_o}{T_w},\qquad \Delta\tau=\frac{1}{T_w}.
  \label{supp:eq:s1}
\end{equation}

We set $T_w=50$ across all benchmarks. By default, visual observations are updated every 50 control steps. At each observation update, the local evolution time $\tau$ is reset to zero, while the action state is carried forward without reinitialization. Specifically, the final action generated in the preceding window serves as the initial action of the next window. Thus, only the local evolution time is reset, whereas the recursive action state is preserved across observation windows.

Training demonstrations consist of discrete action sequences. We perform linear interpolation between adjacent demonstrated actions along $\tau$ to obtain $\gamma(\tau)$. Within each interpolation segment, $\dot\gamma(\tau)=\mathrm d\gamma(\tau)/\mathrm d\tau$ is the corresponding piecewise evolution rate; derivatives at discrete demonstration nodes are assigned according to the convention for adjacent interpolation segments. During training, we sample $\tau\sim\mathcal U[0,1]$ and $\epsilon\sim\mathcal N(0,I)$, and set

\begin{equation}
a=\gamma(\tau)+\sigma\epsilon,
\qquad
F^*(a,\tau)=\dot\gamma(\tau)+k\bigl(\gamma(\tau)-a\bigr).
  \label{supp:eq:s2}
\end{equation}

Neighborhood perturbations are used only to construct supervised query positions, rather than as starting points for generative denoising. Optimization uses the direct transition-field supervision in Eq.~(\ref{eq:training-objective}) of the main paper, and training itself does not require backpropagation through the numerical ODE solution process.

\textbf{Euler Integration, Observation Updates, and Action Anchors.} During execution, we query the action head at the current action $a_n$ and use the forward Euler method to compute

\begin{equation}
a_{n+1}=a_n+\Delta\tau F_\theta(a_n,\tau_n\mid c).
  \label{supp:eq:s3}
\end{equation}

Each integration step corresponds to one control step, and the generated $a_{n+1}$ becomes the action starting point for the next query.

When a new visual observation becomes available, the backbone updates the conditioning representation $c$ and continues integrating from the current action along the action transition field under the updated condition. Action integration and visual-observation updates follow the control frequency and observation schedule, respectively.

\textbf{Global Parameter Configuration.} ODEAction's neighborhood scale and correction gain control the extent of supervision coverage around demonstrations and the strength of recovery for deviating actions, respectively. We perform a grid search over candidate configurations with $k\in[1,10]$ and $\sigma\in[0.01,1.0]$, and select $\sigma=0.1$ and $k=5$. The full method uses fixed $\sigma$ and $k$ throughout the entire evolution interval. Except for dedicated ablation experiments with $\sigma=0$ or $k=0$, this parameter configuration is used across all simulation benchmarks, base policies, and real-robot experiments in this paper.

\section{Experimental Configurations and Evaluation Protocols}

\label{supp:section:b}

\textbf{Representative RoboTwin Tasks.} Controlled comparisons on RoboTwin~\cite{chen2025robotwin} use five representative tasks: Adjust Bottle, Handover Block, Open Laptop, Lift Pot, and Place Can Basket. These cover pose-constrained fine adjustment with a single arm, bimanual handover, articulated-object manipulation, synchronized bimanual transport, and alternating bimanual manipulation, respectively. The five tasks span different motion constraints, bimanual coordination patterns, and end-effector manipulation precision requirements, and are used for the controlled comparisons and analyses in Secs.~\ref{sec:continuation-comparison} and~\ref{sec:ablation} of the main paper and Supplementary Materials~\ref{supp:section:f} and~\ref{supp:section:g}.

\textbf{Action Generation and Execution Configurations.} To compare different action generation methods under the same execution conditions, we keep the vision-language backbone, task-training data, action representation, control frequency, and visual-observation update times identical within each matched comparison. For action-chunking policies, we distinguish the length $H$ of an action chunk predicted in a single pass, the number of control steps actually executed $K_{\mathrm{exec}}$, and the visual-observation update interval $K_{\mathrm{obs}}$. RTC~\cite{black2025real} and Legato~\cite{liu2026learning} retain the action-chunk generation and execution interfaces of their corresponding base policies, introducing cross-chunk continuation mechanisms during inference and training, respectively. In ODEAction, the action head outputs a single-step action evolution rate, and control commands are generated through forward integration. All compared methods execute on matched control timelines and update visual observations at the same control times.

\textbf{Real-Robot Platform and Demonstration Data.} Experiments use an AgileX Cobot Magic 1.0 dual-arm robot equipped with Intel RealSense D435 and Orbbec DaBai RGB-D cameras. Model training and inference use an NVIDIA H100 GPU. Through teleoperation, we collect 100 demonstration trajectories for each of four tasks---Put Bananas, Open Drawer, Place Cup, and Insert Spoon---for a total of 400 trajectories. Each demonstration contains visual observations, robot proprioceptive information, and control actions aligned with a language instruction. The four compared methods are trained on the same task data.

\begin{figure}[htbp]
  \centering
  \includegraphics[width=\linewidth]{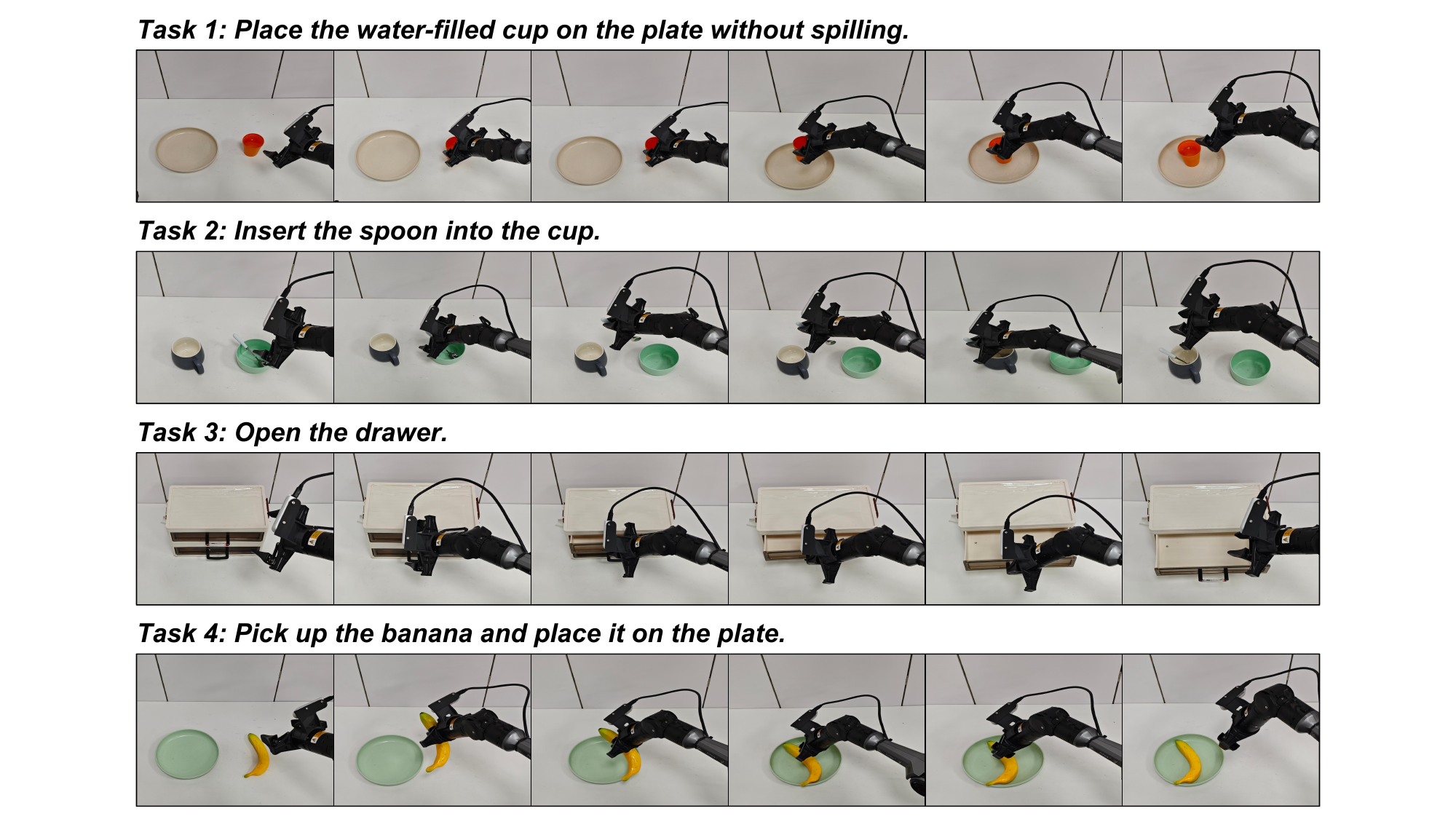}
  \caption{\textbf{Representative execution sequences for real-robot tasks.} Each row corresponds to one task, with frames showing key stages of execution from left to right.}
  \label{supp:fig:s1}
\end{figure}

\textbf{Real-Robot Evaluation Protocol.} Using $\pi_{0.5}$~\cite{black2025pi_} as the common base policy, we evaluate Base Policy, Base Policy + RTC, Base Policy + Legato, and Base Policy + ODEAction. Each method is tested 20 times on each task, yielding a total of 320 real-robot evaluations across the four methods and four tasks. All tasks use binary success criteria: Put Bananas requires correctly placing the bananas on the plate; Open Drawer requires fully opening the designated drawer; Place Cup requires correctly placing a water-filled cup on the plate without spilling water throughout execution; and Insert Spoon requires precisely inserting the spoon into the cup. Task success rates are calculated across all evaluation episodes, while the four control-command trajectory-quality metrics are computed only from the successful episodes of each method; detailed definitions are provided in Supplementary Material~\ref{supp:section:c}. Representative executions of the four tasks are shown in Fig.~\ref{supp:fig:s1}.

\section{Evaluation Metrics}

\label{supp:section:c}

\textbf{Measurement Targets and Sample Scope.} Success rates are calculated across all evaluation episodes. Following the policy-output trajectory evaluation practice used in Legato~\cite{liu2026learning}, NLDLJ, NSPARC, EE-MAJ, and Boundary Jump are computed only on \textbf{successful episodes of each method}, using the control-command sequences directly generated by each action head, rather than high-frequency interpolations from the low-level controller or filtered robot-execution measurements. Success rates and the four trajectory metrics characterize task completion capability and command behavior during successful execution, respectively. Offline rollout errors in Sec.~\ref{sec:ablation} of the main paper and Supplementary Material~\ref{supp:section:g} use common demonstration-trajectory initial conditions, fixed conditioning, and evolution-time grids, and are evaluated independently of online evaluation episodes. Cross-task results are averaged over the tasks listed in each table.

NLDLJ, NSPARC, and EE-MAJ measure the \textbf{Motion Smoothness} of action trajectories. \textbf{Boundary Jump} measures \textbf{Trajectory Continuity} at the action-chunk boundary times of Base Policy; offline rollout error and the ratio of second-half to first-half error characterize the degree of error accumulation during recursive generation.

\textbf{Negative Log Dimensionless Jerk (NLDLJ).} NLDLJ characterizes the higher-order smoothness of an entire action trajectory in the time domain~\cite{Balasubramanian2015OnTA}. Jerk is the third temporal derivative of the action trajectory on the control-step timeline, i.e., the rate of change of acceleration. For an action trajectory of duration $T$, let $v(t)$ denote its scalar speed, $j(t)$ its jerk, and $v_{\mathrm{peak}}=\max_t |v(t)|$ its peak speed. NLDLJ is defined as

\begin{equation}
\mathrm{NLDLJ}
=
\log
\left(
\frac{T^5}{v_{\mathrm{peak}}^2}
\int_0^T
\|j(t)\|_2^2\,dt
\right).
  \label{supp:eq:s4}
\end{equation}

A lower NLDLJ indicates weaker higher-order changes and smoother overall motion. In practice, we compute NLDLJ on joint-action sequences. Gripper actions are excluded because they typically contain discrete opening and closing transitions and do not directly reflect the smoothness of the robotic arm's motion trajectory.

\textbf{Negative Spectral Arc Length (NSPARC).} NSPARC characterizes trajectory smoothness in the frequency domain and is sensitive to oscillations, repeated corrections, and high-frequency velocity changes~\cite{Balasubramanian2015OnTA}. We first obtain a scalar speed sequence from the Euclidean norms of difference vectors between adjacent actions in the joint-action sequence, and then compute the negative spectral arc length of this speed sequence. Gripper dimensions are also excluded. Given a speed signal $v(t)$ with spectral magnitude $|V(\omega)|$, the normalized spectrum is defined as

\begin{equation}
\widehat{V}(\omega)
=
\frac{|V(\omega)|}{|V(0)|}.
  \label{supp:eq:s5}
\end{equation}

The cutoff frequency is adaptively defined as the highest frequency at which the normalized spectral magnitude remains above 0.05. Denote this cutoff frequency by $\omega_c$, and let $\tilde{\omega}=\omega/\omega_c$. NSPARC is defined as

\begin{equation}
\mathrm{NSPARC}(v)
=
\int_0^{\omega_c}
\left[
\left(
\frac{\mathrm d\tilde{\omega}}{\mathrm d\omega}
\right)^2
+
\left(
\frac{\mathrm d\widehat{V}(\omega)}{\mathrm d\omega}
\right)^2
\right]^{1/2}
\mathrm d\omega.
  \label{supp:eq:s6}
\end{equation}

A lower NSPARC indicates that velocity changes are more concentrated in the low-frequency region, with fewer oscillations and high-frequency corrections.

\textbf{End-Effector Mean Absolute Jerk (EE-MAJ).} EE-MAJ directly measures the mean absolute jerk of the end-effector trajectory, characterizing the absolute magnitude of abrupt action changes over the entire trajectory. Let $x(t)$ denote the end-effector position trajectory. Then

\begin{equation}
\mathrm{EE\text{-}MAJ}
=
\frac{1}{T}
\int_0^T
\left\|
\frac{\mathrm d^3x(t)}{\mathrm dt^3}
\right\|_2
\mathrm dt.
  \label{supp:eq:s7}
\end{equation}

For systems with multiple end effectors, we compute EE-MAJ separately for each end effector and then average the values. A lower EE-MAJ indicates fewer abrupt changes in the end-effector trajectory.

\textbf{Boundary Jump.} Boundary Jump measures the discontinuity of control commands when an action-chunking policy switches between adjacent action chunks. Let $B_{\mathrm{Base}}$ denote the set of action-chunk boundary times of Base Policy, and $a_t$ the action output at time $t$. Then

\begin{equation}
J^{\mathrm{bnd}}
=
\frac{1}{|B_{\mathrm{Base}}|}
\sum_{t\in B_{\mathrm{Base}}}
\|a_t-a_{t-1}\|_2.
  \label{supp:eq:s8}
\end{equation}

A lower value indicates more continuous action transitions at chunk boundaries. Because ODEAction does not contain action chunks, its Boundary Jump is computed at the same action-chunk boundary times as the corresponding Base Policy, ensuring that the two methods are compared at the same execution times. RTC, Legato, and direct action-by-action prediction also use the boundary times of the original Base Policy in the matched comparison as common reference times; these boundary times are determined by the actual execution schedule of the corresponding Base Policy.

\textbf{Offline Rollout Error and Error Accumulation Ratio.} Offline rollout error characterizes the deviation of the learned action transition field from demonstration trajectories during recursive generation. Given a demonstration trajectory $\gamma$, we initialize the rollout at its starting point, recursively generate actions using ODEAction's action transition field under fixed conditioning, and compare each generated action with the demonstrated action at the same evolution time. The full-rollout error is defined as

\begin{equation}
E
=
\mathbb{E}
\left[
\frac{1}{Md}
\sum_{n=1}^{M}
\left\|
\hat{a}_n-\gamma(n\Delta\tau)
\right\|_2^2
\right],
  \label{supp:eq:s9}
\end{equation}

where $\hat{a}_n$ is the $n$-th recursively generated action, $M$ is the number of rollout steps, and $d$ is the action dimensionality, consistent with the definition in Supplementary Material~\ref{supp:section:a}; the expectation is taken over the demonstration trajectories used for offline evaluation. We further compute the same type of error over the first and second halves of evolution time, denoted by $E_1$ and $E_2$, respectively, and use $E_2/E_1$ as the \textbf{Error Accumulation Ratio}. A ratio greater than 1 indicates that errors are further amplified in the second half of the rollout; a ratio close to or below 1 indicates that recursive error accumulation is effectively suppressed.

\section{Statistical Association Between BDR and Success-Rate Gains}

\label{supp:section:d}

To quantitatively examine the relationship between \textbf{Boundary Discontinuity Reduction (BDR)} and task-performance gains in Sec.~\ref{sec:chunking-comparison} of the main paper, we use the four LIBERO suites and six EBench tasks as a total of 10 evaluation units. BDR is computed according to Eq.~(\ref{eq:bdr}) of the main paper. For the $i$-th evaluation unit, the success-rate gain, expressed in percentage points, is defined as

\begin{equation}
\Delta \mathrm{SR}_i
=
\mathrm{SR}^{\mathrm{ODEAction}}_i
-
\mathrm{SR}^{\mathrm{Base}}_i.
  \label{supp:eq:s10}
\end{equation}

Based on the evaluation-unit results reported in Table~\ref{tab:main-results} of the main paper, we perform a two-sided Spearman rank correlation test between BDR and success-rate gains, obtaining

\begin{equation}
\rho_s=0.939,
\qquad
p=5.61\times10^{-5}.
  \label{supp:eq:s11}
\end{equation}

The results show a statistically significant strong positive correlation between BDR and success-rate gains: evaluation units with greater reductions in boundary discontinuity generally also achieve larger task-performance gains. This statistical result supports the observation in Sec.~\ref{sec:chunking-comparison} of the main paper that improvements in trajectory continuity and gains in task performance exhibit a consistent positive association across tasks.

\section{Per-Task Results on RoboTwin and Real Robots}

\label{supp:section:e}

To supplement the aggregate RoboTwin results in Table~\ref{tab:generality} of the main paper, Table~\ref{supp:tab:s1} reports the per-task success rates of \textbf{Base Policy + ODEAction} on all 50 tasks under the Clean and Random evaluation settings. The average success rates are \textbf{89.44\%} and \textbf{89.84\%}, respectively, corresponding to the rounded values of \textbf{89.4\%} and \textbf{89.8\%} in the main paper.

\begin{table}[htbp]
  \caption{\textbf{Per-task success rates (\%) of Base Policy + ODEAction on all 50 RoboTwin tasks.} C and R denote the Clean and Random evaluation settings, respectively.}
  \label{supp:tab:s1}
  \centering
  \small
  \setlength{\tabcolsep}{0pt}
  \begin{tabularx}{0.94\linewidth}{@{}X@{\hspace{6pt}}>{\raggedleft\arraybackslash}p{9mm}@{\hspace{8pt}}>{\raggedleft\arraybackslash}p{9mm}@{\hspace{20pt}}X@{\hspace{6pt}}>{\raggedleft\arraybackslash}p{9mm}@{\hspace{8pt}}>{\raggedleft\arraybackslash}p{9mm}@{}}
    \toprule
    Task & C & R & Task & C & R \\
    \midrule
    Adjust Bottle & 100 & 100 & Place Can Basket & 74 & 74 \\
    Beat Block Hammer & 95 & 94 & Place Cans Plastic Box & 100 & 99 \\
    Blocks Ranking RGB & 99 & 98 & Place Container Plate & 99 & 99 \\
    Blocks Ranking Size & 85 & 86 & Place Dual Shoes & 89 & 88 \\
    Click Alarm Clock & 73 & 77 & Place Empty Cup & 97 & 97 \\
    Click Bell & 84 & 87 & Place Fan & 91 & 92 \\
    Dump Bin Big Bin & 100 & 100 & Place Mouse Pad & 73 & 79 \\
    Grab Roller & 100 & 100 & Place Object Basket & 87 & 88 \\
    Handover Block & 94 & 91 & Place Object Scale & 89 & 89 \\
    Handover Mic & 100 & 99 & Place Object Stand & 99 & 98 \\
    Hanging Mug & 55 & 52 & Place Phone Stand & 71 & 75 \\
    Lift Pot & 100 & 100 & Place Shoe & 96 & 99 \\
    Move Can Pot & 80 & 79 & Press Stapler & 100 & 98 \\
    Move Pill Bottle Pad & 98 & 99 & Put Bottles Dustbin & 87 & 83 \\
    Move Playing Card Away & 98 & 97 & Put Object Cabinet & 93 & 94 \\
    Move Stapler Pad & 56 & 70 & Rotate QR Code & 88 & 89 \\
    Open Laptop & 99 & 100 & Scan Object & 92 & 90 \\
    Open Microwave & 97 & 100 & Shake Bottle & 100 & 100 \\
    Pick Diverse Bottles & 85 & 84 & Shake Bottle Horizontally & 100 & 100 \\
    Pick Dual Bottles & 88 & 89 & Stack Blocks Three & 96 & 91 \\
    Place A2B Left & 90 & 94 & Stack Blocks Two & 100 & 100 \\
    Place A2B Right & 87 & 92 & Stack Bowls Three & 82 & 79 \\
    Place Bread Basket & 89 & 87 & Stack Bowls Two & 99 & 100 \\
    Place Bread Skillet & 95 & 90 & Stamp Seal & 67 & 71 \\
    Place Burger Fries & 99 & 99 & Turn Switch & 57 & 56 \\
    \midrule
    \multicolumn{4}{@{}l}{\textbf{Average}} & \textbf{89.44} & \textbf{89.84} \\
    \bottomrule
  \end{tabularx}
\end{table}

\textbf{Real-Robot Per-Task Results.} Figure~\ref{fig:real-tasks}(c--d) of the main paper shows the average success rate and trajectory quality of each method across the four real-robot tasks. Table~\ref{supp:tab:s2} further provides the complete quantitative results for the four methods on each task, including success rate, NLDLJ, NSPARC, EE-MAJ, and Boundary Jump. The experimental platform, success criteria, and evaluation protocol are described in Supplementary Material~\ref{supp:section:b}, and the calculation of trajectory-quality metrics is provided in Supplementary Material~\ref{supp:section:c}.

\begin{table}[htbp]
  \caption{\textbf{Comparison of success rates and control-command trajectory quality for four methods on four real-robot tasks.} With $\pi_{0.5}$ as the base policy, values in each cell correspond to \textbf{Base Policy / +RTC / +Legato / +ODEAction}, in that order. Higher SR is better, while lower values are better for the other four metrics; EE-MAJ is reported in units of $10^{-3}$, and bold indicates the best results. Success rates are calculated from 20 trials per method per task, while trajectory-quality metrics are computed only from the successful episodes of each method.}
  \label{supp:tab:s2}
  \centering
  \small
  \setlength{\tabcolsep}{4pt}
  \begin{tabular}{lccccc}
    \toprule
    Task & SR (\%) $\uparrow$ & NLDLJ $\downarrow$ & NSPARC $\downarrow$ & EE-MAJ $\downarrow$ & Boundary Jump $\downarrow$ \\
    \midrule
    Put Bananas & \shortstack[c]{70.00 / 60.00 /\\70.00 / \textbf{75.00}} & \shortstack[c]{25.84 / 25.38 /\\25.17 / \textbf{24.66}} & \shortstack[c]{2.757 / 2.691 /\\2.656 / \textbf{2.595}} & \shortstack[c]{2.62 / 2.49 /\\2.42 / \textbf{1.79}} & \shortstack[c]{0.1562 / 0.1519 /\\0.1532 / \textbf{0.1474}} \\
    Open Drawer & \shortstack[c]{95.00 / 95.00 /\\90.00 / \textbf{100.00}} & \shortstack[c]{24.98 / 24.17 /\\24.34 / \textbf{22.47}} & \shortstack[c]{2.873 / 2.814 /\\2.829 / \textbf{2.734}} & \shortstack[c]{3.24 / 2.87 /\\2.95 / \textbf{2.33}} & \shortstack[c]{0.1147 / 0.1108 /\\0.1089 / \textbf{0.1027}} \\
    Place Cup & \shortstack[c]{70.00 / 65.00 /\\65.00 / \textbf{90.00}} & \shortstack[c]{24.86 / 24.25 /\\23.99 / \textbf{23.01}} & \shortstack[c]{2.365 / 2.341 /\\2.318 / \textbf{2.296}} & \shortstack[c]{3.06 / 2.77 /\\2.56 / \textbf{1.42}} & \shortstack[c]{0.1517 / 0.1434 /\\0.1411 / \textbf{0.1374}} \\
    Insert Spoon & \shortstack[c]{25.00 / 35.00 /\\40.00 / \textbf{55.00}} & \shortstack[c]{25.03 / 24.74 /\\24.77 / \textbf{23.92}} & \shortstack[c]{2.146 / 2.125 /\\2.130 / \textbf{2.097}} & \shortstack[c]{3.13 / 2.74 /\\2.69 / \textbf{1.68}} & \shortstack[c]{0.1455 / 0.1406 /\\0.1419 / \textbf{0.1365}} \\
    \midrule
    \textbf{Average} & \shortstack[c]{65.00 / 63.75 /\\66.25 / \textbf{80.00}} & \shortstack[c]{25.18 / 24.64 /\\24.57 / \textbf{23.52}} & \shortstack[c]{2.535 / 2.493 /\\2.483 / \textbf{2.431}} & \shortstack[c]{3.01 / 2.72 /\\2.66 / \textbf{1.81}} & \shortstack[c]{0.1420 / 0.1367 /\\0.1363 / \textbf{0.1310}} \\
    \bottomrule
  \end{tabular}
\end{table}

\section{Direct Next-Action Prediction and Continuous Action Evolution}

\label{supp:section:f}

ODEAction transforms action generation from future action-chunk prediction into continuous evolution starting from the current action. However, simply removing action chunks and establishing recursive dependencies between adjacent actions does not imply that stable action evolution dynamics have been learned. To examine whether ODEAction's benefits arise solely from this chunk-free recursive formulation, we compare it with Direct Next-Action prediction (Direct Next-Action). Both methods generate subsequent actions step by step, but use different learning objectives and generation mechanisms.

\textbf{Direct Next-Action Prediction (Direct Next-Action).} Given the current action, evolution time, and vision-language conditioning, Direct Next-Action directly regresses the next action value, with the recursion

\begin{equation}
\hat a_{n+1}=G_\psi(\hat a_n,\tau_n\mid c),
  \label{supp:eq:s12}
\end{equation}

where $\psi$ denotes the parameters of the prediction network $G_\psi$. Supervised by adjacent action pairs from demonstrations, this network learns a direct mapping from the current action to the next action and uses the predicted action as the input to the next step during inference. In contrast, ODEAction learns an action evolution rate jointly defined by the demonstration tangent and the deviation correction term, and advances actions through ODE integration. Its supervision also covers demonstration neighborhoods, providing corresponding evolution directions for off-trajectory actions.

We use the StarVLA-$\pi$ base policy and five representative RoboTwin tasks from Sec.~\ref{sec:continuation-comparison} of the main paper, keeping the vision-language backbone, training data, control frequency, and visual-observation update schedule identical. Task performance is measured by success rate, and control-command trajectory quality is measured by NLDLJ, NSPARC, EE-MAJ, and Boundary Jump, following the same statistical protocol as Sec.~\ref{sec:experimental-setup} of the main paper.

\begin{table}[htbp]
  \caption{\textbf{Task performance and control-command trajectory quality of Direct Next-Action and ODEAction on five representative RoboTwin tasks.} Values in each cell are ordered as \textbf{Direct Next-Action / ODEAction}, and bold indicates the best results. Higher SR is better, while lower values are better for the remaining metrics; EE-MAJ is reported in units of $10^{-3}$.}
  \label{supp:tab:s3}
  \centering
  \small
  \setlength{\tabcolsep}{4pt}
  \begin{tabular}{lccccc}
    \toprule
    Task & SR (\%) $\uparrow$ & NLDLJ $\downarrow$ & NSPARC $\downarrow$ & EE-MAJ $\downarrow$ & Boundary Jump $\downarrow$ \\
    \midrule
    Adjust Bottle & 92.0 / \textbf{94.0} & 32.87 / \textbf{21.54} & 3.891 / \textbf{1.991} & 5.614 / \textbf{0.625} & 0.2914 / \textbf{0.1507} \\
    Handover Block & 67.0 / \textbf{82.0} & 37.26 / \textbf{26.67} & 4.622 / \textbf{2.735} & 8.327 / \textbf{0.581} & 0.2731 / \textbf{0.0391} \\
    Open Laptop & 83.0 / \textbf{87.0} & 34.95 / \textbf{22.73} & 4.107 / \textbf{1.848} & 6.923 / \textbf{0.449} & 0.3236 / \textbf{0.0264} \\
    Lift Pot & 88.0 / \textbf{91.0} & 33.58 / \textbf{20.69} & 3.673 / \textbf{1.566} & 8.541 / \textbf{0.511} & 0.4338 / \textbf{0.1613} \\
    Place Can Basket & 62.0 / \textbf{68.0} & 35.76 / \textbf{23.18} & 4.414 / \textbf{2.338} & 6.350 / \textbf{0.500} & 0.2995 / \textbf{0.0462} \\
    \midrule
    \textbf{Average} & 78.4 / \textbf{84.4} & 34.88 / \textbf{22.96} & 4.141 / \textbf{2.096} & 7.151 / \textbf{0.533} & 0.3243 / \textbf{0.0847} \\
    \bottomrule
  \end{tabular}
\end{table}

\textbf{Results.} As shown in Table~\ref{supp:tab:s3}, Direct Next-Action achieves an average success rate of 78.4\%, indicating that direct action-by-action prediction can complete a substantial portion of manipulation tasks, but the trajectory quality of its generated actions is markedly worse than that of ODEAction. ODEAction raises the average success rate to 84.4\%, an improvement of 6.0 percentage points, while improving all four trajectory metrics on all five tasks: average NLDLJ and NSPARC decrease from 34.88 and 4.141 to 22.96 and 2.096, respectively; EE-MAJ decreases from 7.151 to 0.533; and Boundary Jump decreases from 0.3243 to 0.0847. These results indicate that, although direct next-action prediction establishes generative dependencies between adjacent actions, its recursive process remains prone to pronounced command jitter and abrupt action changes.

This comparison further clarifies ODEAction's modeling advantage: \textbf{chunk-free action-by-action recursion alone is insufficient to achieve high-quality continuous action generation.} Direct Next-Action directly learns a mapping between adjacent demonstrated actions, whereas ODEAction formulates action generation as a continuous evolution process, characterizes evolution rates by learning an action transition field, and combines neighborhood supervision with deviation correction to provide evolution directions after deviations from demonstrations. Together with the component ablation results in Sec.~\ref{sec:ablation} of the main paper, these findings show that ODEAction's gains in task performance and trajectory quality arise not only from removing action chunks, but are also closely related to continuous action evolution modeling and its corresponding field-supervision design.

\section{Neighborhood Supervision and Recursive Behavior of the Action Transition Field}

\label{supp:section:g}

ODEAction treats the current action as the state of a recursive evolution process. The Action Transition Field specifies how this state changes at each evolution time. During training, query actions are sampled from a fixed-scale neighborhood around the demonstration at every evolution time, providing supervision for off-trajectory states throughout the entire evolution interval. Given a demonstration action curve $\gamma(\tau)$, ODEAction's training-query distribution is

\begin{equation}
q_{\mathrm U}(a\mid\tau,\gamma)
=\mathcal N\!\bigl(\gamma(\tau),\sigma^2I\bigr).
  \label{supp:eq:s13}
\end{equation}

The corresponding target action transition field is

\begin{equation}
F_{\mathrm U}^{*}(a,\tau)
=\dot\gamma(\tau)+k\bigl(\gamma(\tau)-a\bigr).
  \label{supp:eq:s14}
\end{equation}

Here, the demonstration tangent $\dot\gamma(\tau)$ provides the direction of progression along the reference trajectory, while the correction term $k(\gamma(\tau)-a)$ provides proportional correction toward the trajectory for off-trajectory queries. This progression-and-correction principle is an established design in robot control and motion modeling, with related formulations in robot path-following vector fields~\cite{kapitanyuk2017guiding}, stable dynamical systems for demonstrated motions~\cite{li2025movement}, and progression--attraction flows for policy imitation~\cite{li2026geometry}.

By maintaining the same neighborhood scale at every evolution time, ODEAction continuously supervises sampled off-trajectory actions across the entire evolution interval. This provides consistent coverage of the demonstration neighborhood at different stages of action evolution.

Within flow-based action generation, SFP~\cite{jiang2025streaming} is a concrete example of using this established progression-and-correction construction. Its associated conditional action distribution contracts over evolution time as

\begin{equation}
q_{\mathrm S}(a\mid\tau,\gamma)
=\mathcal N\!\left(\gamma(\tau),\sigma_0^2 e^{-2k\tau}I\right),
  \label{supp:eq:s16}
\end{equation}

with
\[
\sigma_{\mathrm S}(\tau)=\sigma_0e^{-k\tau}.
\]

Here, $\sigma_0$ is the initial neighborhood standard deviation, and $\sigma_{\mathrm S}(\tau)$ is the standard deviation at evolution time $\tau$.

By contrast, ODEAction maintains a fixed-scale training-query distribution $q_U$ independently of the correction gain $k$, providing off-trajectory supervision at the same neighborhood scale throughout evolution. At inference, rollout starts directly from the current action.

This sustained neighborhood coverage is particularly important for recursive inference. As rollout proceeds, prediction and execution deviations can move the current action away from the demonstration trajectory. Because the Action Transition Field is supervised on sampled off-trajectory actions at every evolution stage, it learns corresponding evolution and recovery directions throughout the recursive process. This design supports more robust action evolution under deviations and helps suppress the accumulation of recursive errors.

To isolate the effect of the training-query distribution, we construct a Shrinking variant by replacing $q_U$ with the contracting distribution $q_S$, while keeping the target action transition field, StarVLA-$\pi$ base policy, and five representative RoboTwin tasks unchanged. This controlled comparison isolates the effect of fixed versus contracting neighborhood support on learning the recursive action transition field. We evaluate both variants in terms of task success, control-command trajectory quality, and offline recursive error.

\begin{table}[!t]
  \caption{\textbf{Task performance and control-command trajectory quality of Shrinking Construction and Uniform ODEAction on five representative RoboTwin tasks.} Values in each cell are ordered as \textbf{Shrinking / Uniform ODEAction}, and bold indicates the best results. Higher SR is better, while lower values are better for the remaining metrics; EE-MAJ is reported in units of $10^{-3}$. The four trajectory-quality metrics are computed from the successful episodes of each method.}
  \label{supp:tab:s4}
  \centering
  \small
  \setlength{\tabcolsep}{4pt}
  \begin{tabular}{lccccc}
    \toprule
    Task & SR (\%) $\uparrow$ & NLDLJ $\downarrow$ & NSPARC $\downarrow$ & EE-MAJ $\downarrow$ & Boundary Jump $\downarrow$ \\
    \midrule
    Adjust Bottle & 91.0 / \textbf{94.0} & 22.31 / \textbf{21.54} & 2.004 / \textbf{1.991} & 0.813 / \textbf{0.625} & 0.1585 / \textbf{0.1507} \\
    Handover Block & 74.0 / \textbf{82.0} & 28.36 / \textbf{26.67} & 2.865 / \textbf{2.735} & 1.061 / \textbf{0.581} & 0.0594 / \textbf{0.0391} \\
    Open Laptop & 86.0 / \textbf{87.0} & 24.91 / \textbf{22.73} & 1.955 / \textbf{1.848} & 0.923 / \textbf{0.449} & 0.0695 / \textbf{0.0264} \\
    Lift Pot & 89.0 / \textbf{91.0} & 21.78 / \textbf{20.69} & 1.638 / \textbf{1.566} & 0.781 / \textbf{0.511} & 0.1824 / \textbf{0.1613} \\
    Place Can Basket & 53.0 / \textbf{68.0} & 24.71 / \textbf{23.18} & 2.494 / \textbf{2.338} & 0.750 / \textbf{0.500} & 0.0655 / \textbf{0.0462} \\
    \midrule
    \textbf{Average} & 78.6 / \textbf{84.4} & 24.41 / \textbf{22.96} & 2.191 / \textbf{2.096} & 0.866 / \textbf{0.533} & 0.1071 / \textbf{0.0847} \\
    \bottomrule
  \end{tabular}
\end{table}

\begin{table}[!t]
  \caption{\textbf{Offline recursive errors of Shrinking Construction and Uniform ODEAction on five representative RoboTwin tasks.} Values in each cell are ordered as \textbf{Shrinking / Uniform ODEAction}, and bold indicates the best results. $E$, $E_1$, and $E_2$ are reported in units of $10^{-4}$, with lower values being better for all three; a lower $E_2/E_1$ indicates weaker relative error growth in the second half. The ratio in the Average row is obtained by dividing the cross-task average of $E_2$ by the cross-task average of $E_1$.}
  \label{supp:tab:s5}
  \centering
  \small
  \setlength{\tabcolsep}{4pt}
  \begin{tabular}{lcccc}
    \toprule
    Task & $E$ $\downarrow$ & $E_1$ $\downarrow$ & $E_2$ $\downarrow$ & $E_2/E_1$ $\downarrow$ \\
    \midrule
    Adjust Bottle & 5.55 / \textbf{2.90} & 4.40 / \textbf{3.14} & 6.69 / \textbf{2.65} & 1.52 / \textbf{0.84} \\
    Handover Block & 8.64 / \textbf{4.68} & 7.05 / \textbf{5.03} & 10.22 / \textbf{4.32} & 1.45 / \textbf{0.86} \\
    Open Laptop & 7.39 / \textbf{3.79} & 5.35 / \textbf{3.90} & 9.42 / \textbf{3.67} & 1.76 / \textbf{0.94} \\
    Lift Pot & 5.66 / \textbf{3.15} & 4.60 / \textbf{3.38} & 6.72 / \textbf{2.91} & 1.46 / \textbf{0.86} \\
    Place Can Basket & 10.38 / \textbf{6.73} & 7.95 / \textbf{6.65} & 12.80 / \textbf{6.80} & 1.61 / \textbf{1.02} \\
    \midrule
    \textbf{Average} & 7.52 / \textbf{4.25} & 5.87 / \textbf{4.42} & 9.17 / \textbf{4.07} & 1.56 / \textbf{0.92} \\
    \bottomrule
  \end{tabular}
\end{table}

Table~\ref{supp:tab:s4} shows that the time-uniform neighborhood construction achieves better overall performance across the five tasks: the average success rate increases from 78.6\% for Shrinking to 84.4\% for ODEAction, while NLDLJ, NSPARC, EE-MAJ, and Boundary Jump also improve on all five tasks. In particular, success rates on Place Can Basket and Handover Block increase by 15 and 8 percentage points, respectively; average EE-MAJ decreases from 0.866 to 0.533, and Boundary Jump from 0.1071 to 0.0847. Because both constructions use the same local target field, these differences indicate that the distribution of training supervision over evolution time itself affects the task performance and trajectory quality of action generation.

We further use the offline demonstration-trajectory protocol in Supplementary Material~\ref{supp:section:c} to examine the recursive error behavior of the two constructions. Starting from the same demonstration initial conditions, the models recursively generate actions under fixed conditioning and compare them with the demonstrated actions at corresponding evolution times. Table~\ref{supp:tab:s5} reports the full-trajectory error $E$, first-half error $E_1$, second-half error $E_2$, and their ratio $E_2/E_1$. Unlike online command-trajectory metrics, offline errors are computed on a common evaluation set of demonstration trajectories.

Table~\ref{supp:tab:s5} further reveals differences between the two supervision constructions during recursion. The average full-trajectory error is 7.52 for Shrinking and 4.25 for Uniform ODEAction; their first-half errors are 5.87 and 4.42, respectively, with the gap widening further in the second half to 9.17 and 4.07. Shrinking has an error accumulation ratio of 1.56, with second-half error growth on all five tasks; Uniform ODEAction has an error accumulation ratio of 0.92, with second-half errors lower than first-half errors on all tasks except Place Can Basket, whose ratio is close to 1. Together with the results in Table~\ref{supp:tab:s4}, time-uniform neighborhood supervision not only improves task completion capability and control-command trajectory quality, but also exhibits better control of recursive error in the evaluated settings. The contrast is particularly evident in the second half of recursive rollout: the shrinking construction exhibits increasing error, whereas the time-uniform construction reduces the error from the first half to the second half.

\end{document}